\documentclass[sigconf]{acmart}

\AtBeginDocument{}
\setcopyright{acmlicensed}
\copyrightyear{2026}
\acmYear{2026}

\acmDOI{XXXXXXX.XXXXXXX}
\acmConference[MM'26]{ACM Multimedia 2026}{November, 2026}{Rio de Janeiro, Brazil}
\usepackage{booktabs}
\usepackage{multirow}
\usepackage{makecell}
\usepackage{graphicx}
\usepackage{booktabs}
\usepackage{float}
\usepackage{tabularx}
\usepackage{threeparttable}
\usepackage{hyperref}
\usepackage[table]{xcolor}
\renewcommand\footnotetextcopyrightpermission[1]{} 
\begin{document}
\title{EA-Nav: Learning Safe Visual Navigation Policies with Embodiment Awareness}

\author{Jialu Zhang}
\authornote{These authors contributed equally to this paper.}
\affiliation{
  \institution{Zhejiang University}
  \city{Hangzhou}
  \state{Zhejiang}
  \country{China}
}
\email{jialu_z@zju.edu.cn}

\author{Yong Du}
\authornotemark[1]
\affiliation{%
  \institution{Zhejiang University}
  \city{Hangzhou}
  \state{Zhejiang}
  \country{China}
}
\email{yongd@zju.edu.cn}

\author{Xianda Guo}
\authornotemark[1]
\affiliation{%
  \institution{Wuhan University}
  \city{Wuhan}
  \state{Hubei}
  \country{China}
}
\email{xianda_guo@163.com}

\author{Shunwang Sun}
\affiliation{%
  \institution{Zhejiang University}
  \city{Hangzhou}
  \state{Zhejiang}
  \country{China}
}
\email{shunwang_sun@zju.edu.cn}

\author{Xinqi Liu}
\affiliation{%
  \institution{Hunan University}
  \city{Changsha}
  \state{Hunan}
  \country{China}
}
\email{liuxinqi@hnu.edu.cn}

\author{Yue Sun}
\affiliation{%
  \institution{Zhejiang University}
  \city{Hangzhou}
  \state{Zhejiang}
  \country{China}
}
\email{yue_sun@zju.edu.cn}

\author{Guodong Lu}
\affiliation{%
  \institution{Zhejiang University}
  \city{Hangzhou}
  \state{Zhejiang}
  \country{China}
}
\email{lugd@zju.edu.cn}

\author{Wei Sui}
\authornote{Corresponding authors.}
\affiliation{%
  \institution{D-Robotics}
  \city{Beijing}
  \country{China}
}
\email{wei.sui@d-robotics.cc}

\author{Jituo Li}
\authornotemark[2]
\affiliation{%
  \institution{Zhejiang University}
  \city{Hangzhou}
  \state{Zhejiang}
  \country{China}
}
\email{jituo_li@zju.edu.cn}
\renewcommand{\shortauthors}{Zhang et al.}
\begin{abstract}
Cross-embodiment navigation is a key challenge in embodied intelligence. Due to differences in embodiment, the same visual observation may imply different actions for different agents, making prediction ambiguous when relying solely on vision. Existing studies mainly rely on reinforcement learning, which requires large-scale interaction and careful reward design, making it difficult to support scalable pretraining and real-world adaptation. In contrast, imitation-learning-based approaches remain limited. To address these challenges, we propose an imitation-learning-based embodiment-aware navigation framework with a modular multi-stage design. In pretraining, we construct a cross-embodiment navigation dataset from Internet videos and introduce embodiment geometry as conditional tokens to reduce action ambiguity under the same observation. In fine-tuning, we design a multimodal information injection mechanism based on a decoupled architecture. Specifically, we design a trajectory augmentation strategy to generate high-risk samples, which are used to train spatial perception and risk-aware correction separately, thereby explicitly incorporating embodiment geometry for safe navigation. Experimental results show that the proposed method effectively improves navigation performance across different embodiment settings, demonstrating the effectiveness of incorporating embodiment geometry into embodied navigation. Project page: \url{https://ea-nav.github.io/}.
\end{abstract}


\begin{teaserfigure}
  \includegraphics[width=\textwidth]{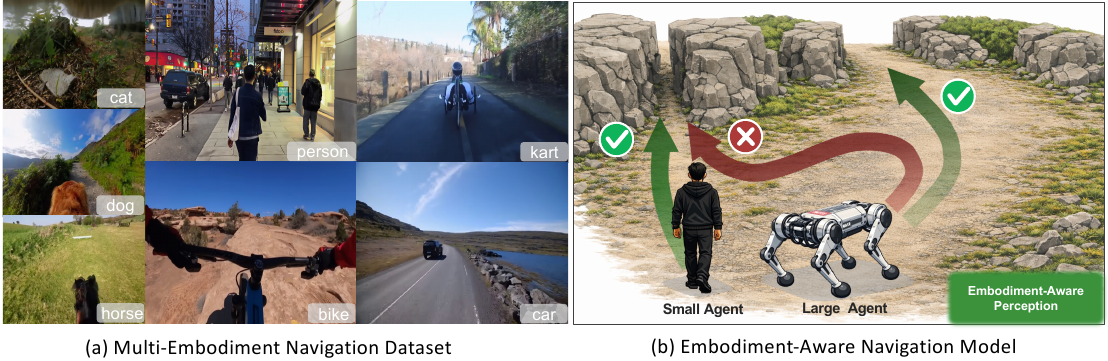}
  \caption{Overview of our contributions. (a) We collect large-scale first-person videos from the Internet across diverse embodiments (e.g., humans, animals, and vehicles), and build a cross-embodiment navigation dataset through annotation. (b) We introduce an embodiment-geometry modality injection mechanism, allowing the navigation model to predict safe trajectories conditioned on embodiment geometry.}
  \Description{fig1}
  \label{fig:1}
\end{teaserfigure}
\maketitle

\section{Introduction}
In recent years, vision-based navigation has achieved significant progress in embodied intelligence~\cite{wu2025aeroduo,wang2025trackvla,wang2025rethinking,zhang2025embodied,hirose2025omnivla}. As a key component, the navigation model is responsible for converting high-level goals into low-level executable trajectories for the robot. However, most existing studies~\cite{shah2023gnm,sridhar2024nomad, cai2025navdp,hirose2025omnivla} primarily rely on visual observations alone, while overlooking the importance of embodiment information, especially geometry and physical morphology, as a critical modality. Since different embodiments vary substantially in size and traversability, relying on a single visual modality prevents the model from predicting reliable trajectories conditioned on the robot’s actual geometric characteristics, thereby becoming a bottleneck for embodied navigation performance.

A straightforward way to incorporate embodiment information is to employ reinforcement learning (RL) to model the interaction between the embodiment and the environment through trial-and-error exploration~\cite{wang2025x,eftekhar2024one,cheng2024navila}. However, RL suffers from a large exploration space and unstable convergence, and thus requires imitation learning (IL) to provide a well-initialized pretrained model~\cite{chen2025socialnav, yang2025nav}. In this work, we focus on how to effectively inject embodiment geometry into the IL stage. Nevertheless, the existing IL-based method~\cite{hirose2023exaug} for cross-embodiment navigation faces two major challenges. First, it is difficult to obtain sufficient training data that reflects how different embodiments interact with the environment. Second, expert demonstrations are typically discretized into waypoint sequences, which breaks the mapping between the continuous physical action space and the visual environment. As a result, when a trajectory changes, it becomes difficult to determine whether the modified trajectory remains safe in the current environment.

To address these challenges, we propose an embodiment-aware navigation framework. In the pretraining stage, we construct a heterogeneous navigation dataset from Internet videos~\cite{liu2025citywalker,hirose2025lelan,chen2026imaginav}, covering multiple embodiment categories such as humans, animals, and vehicles. Since such unstructured data, as illustrated on the left side of Figure~\ref{fig:1}, cannot be directly represented in a low-dimensional geometric space, we introduce embodiment geometry as a conditional token. This allows the navigation model to acquire a preliminary sense of embodiment awareness. For example, cats tend to pass through narrow spaces, whereas cars are more likely to detour around them. By conditioning the model on embodiment geometry, our framework effectively alleviates the ambiguity of action prediction under the same visual observation. Notably, this differs from diffusion-based methods for multimodal action prediction~\cite{chi2025diffusion,chen2026imaginav}, which explicitly model multiple plausible actions under the same observation. In contrast, by conditioning on embodiment geometry, our model resolves such ambiguity across embodiments and learns more stable embodiment-specific action predictions.

In the fine-tuning stage, we further explicitly inject embodiment geometry into the navigation model using a small amount of high-quality navigation data. Unlike the prior approach~\cite{hirose2023exaug} that directly imposes constraints on discrete waypoints, we focus on the fact that discrete trajectory representations fail to capture collision risks in continuous space. To overcome this issue, we adjust the entire trajectory in continuous space rather than operating on individual waypoints. This design is particularly suitable for our short-horizon prediction setting, where an overall rotation of the predicted trajectory is sufficient to achieve effective correction. However, not all trajectories require additional adjustment. For example, when a small embodiment moves in an open space, extra correction is usually unnecessary. Motivated by this observation, we design a \emph{Spatial Perception} module to predict the minimum distance between the trajectory and surrounding obstacles and to identify high-risk trajectories. The \emph{Risk-Aware Correction} module is triggered only when the predicted risk exceeds a predefined safety threshold. Furthermore, because real navigation datasets provide limited high-risk trajectories and corresponding correction samples, we design a risk trajectory augmentation strategy to generate augmented training samples for effective learning of embodiment-aware Spatial Perception and Risk-Aware Correction during fine-tuning.

In summary, we propose an IL-based method for incorporating embodiment information into navigation models, enabling the model to infer safer trajectories conditioned on the robot’s geometric properties. Compared with navigation models using only visual modality, our method explicitly captures the influence of embodiment geometry on trajectory prediction. The main contributions of this work are summarized as follows:
\begin{enumerate}
    \item We propose an IL-based framework for cross-embodiment navigation, which explicitly incorporates embodiment geometry into the navigation model, enabling it to learn embodiment-aware action decisions under the same visual observation rather than relying on vision alone.
    \item We develop a two-stage embodiment information injection strategy that introduces embodiment geometry as conditional tokens in pretraining and explicitly incorporates it into fine-tuning for geometry-aware navigation.
    \item To support effective training, we construct a heterogeneous cross-embodiment navigation dataset and design trajectory augmentation with decoupled training for the distance perception and trajectory correction modules, addressing the scarcity of real high-risk samples.
\end{enumerate}

\section{Related Works}
\subsection{Vision-Based Navigation}
Vision-based navigation has achieved significant progress in recent years with the development of deep learning~\cite{shah2023gnm,sridhar2024nomad,cai2025navdp,liu2025citywalker,bar2025navigation,jing2024two}. Early methods~\cite{shah2023gnm, shah2023vint} typically employed the deterministic model to predict future trajectories based on visual observations and goal information. Subsequently, the diffusion model~\cite{chi2025diffusion} were introduced to model multimodal action distributions and alleviate regression-to-the-mean issues~\cite{sridhar2024nomad, cai2025navdp}. Beyond architectural improvements, recent works have also incorporated priors from vision foundation models into navigation tasks, such as depth estimation~\cite{nayak2025metricnet}, visual encoders~\cite{liu2025citywalker}, future observation prediction~\cite{bar2025navigation,nie2025wmnav}, and open-vocabulary spatial reasoning~\cite{ortega2025div}, to enhance spatial perception and long-horizon planning capabilities~\cite{nayak2025metricnet,liu2025citywalker,cheng2024navila,shah2026wildos}. Despite these advances, most existing methods still focus on the visual modality and lack explicit modeling of embodiment geometry, which limits their adaptability in cross-embodiment scenarios.

\subsection{Embodiment-Aware Navigation}
Existing research on embodiment-aware visual navigation remains limited, particularly within the imitation learning (IL) framework~\cite{zhang2025embodied,xi2026cerlp}. A common approach is to train reinforcement learning policies in simulation by allowing robots with different embodiment sizes to interact with the environment~\cite{eftekhar2024one}. However, such methods often suffer from a large exploration space and unstable convergence. To alleviate this issue, a recent work~\cite{wang2025x} collected navigation data for different embodiments in simulated environments and trained navigation models using IL. Nevertheless, this approach still relies on simulated data, making it difficult to bridge the sim-to-real gap in real-world deployment. Therefore, there remains a need for an embodiment-aware navigation method within the IL framework~\cite{xi2026cerlp,zhu2026sysnav} that can support both large-scale pretraining and real-world fine-tuning.

Within the IL framework, ExAug~\cite{hirose2023exaug} proposes an embodiment-aware training strategy based on signed distance fields, providing useful insights for subsequent research. However, this method relies on discrete waypoints to establish the relationship between continuous motion and the environment, which limits its ability to accurately reflect trajectory safety in continuous space. Moreover, it adjusts future trajectories based on feedback at the current timestep, which may lead to short-sighted navigation decisions. Overall, existing methods still lack an effective way to model embodiment information within the IL framework.

\subsection{Navigation Datasets}
Existing visual navigation datasets can be broadly divided into three categories: real-world datasets, synthetic datasets, and pseudo-labeled datasets. Real-world datasets, such as GND~\cite{liang2025gnd}, SiT~\cite{bae2023sit}, THUD~\cite{tang2024mobile} and SCAND~\cite{karnan2022socially}, are typically collected using specific robotic platforms, while they  are often limited in scene diversity or embodiment coverage. Synthetic datasets, including NavDP~\cite{cai2025navdp} and TartanGround~\cite{patel2025tartanground}, offer better scalability and support multiple viewpoints or platforms, yet still rely on simulated environments~\cite{zheng2025dvs}. More recently, pseudo-labeled datasets such as CityWalker~\cite{liu2025citywalker} and FrodoBots-2k~\cite{hirose2025learning} have leveraged large-scale internet videos or globally collected robot data to automatically provide supervision signals.
Nevertheless, existing datasets still lack coverage across multiple embodiment types. To fill this gap, we construct a heterogeneous dataset from internet videos spanning diverse embodiments, environments, and behavioral patterns, and further generate pseudo-labels using Depth-Anything-3~\cite{lin2025depth} for training.


\section{Method}
\subsection{Preliminaries}
This paper aims to address the navigation problem for agents with diverse embodiment parameters. To achieve this, we explicitly introduce the embodiment geometry of the robot, denoted as $\mathbf{m}$, as a conditional variable into the policy network $\pi_\theta$. Specifically, the embodiment parameter $\mathbf{m} \in \mathbb{R}^4$ is defined as :
\begin{equation}
    \mathbf{m} = [L_b, W_b, H_b, P_{\max}]^\top,
\end{equation}
where $L_b$, $W_b$, and $H_b$ represent the length, width, and height of the agent, respectively, and $P_{\max}$ denotes maximum traversable height. Based on this geometric attribute vector, the policy network is able to learn embodiment-conditioned navigation behaviors.

\begin{figure*}[t]
  \centering
  \includegraphics[width=\textwidth]{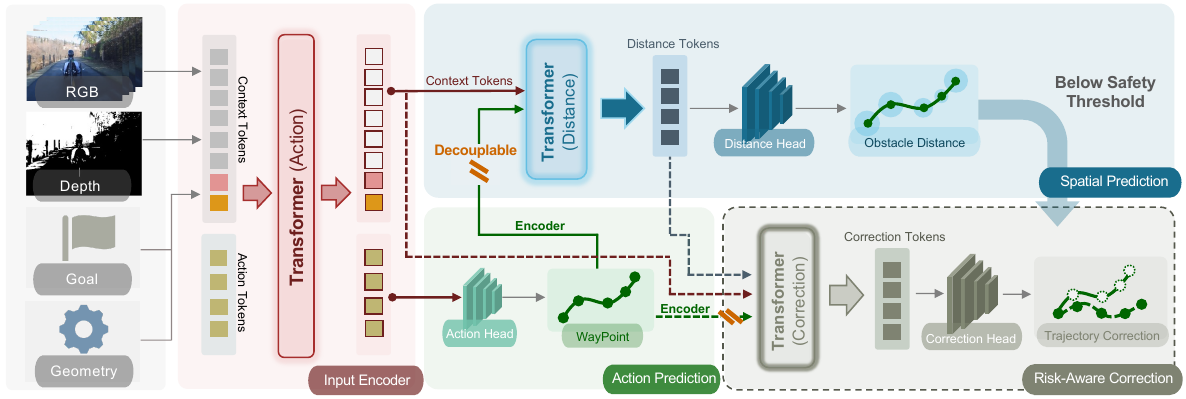}
\caption{Overview of the EA-Nav framework. The Input Encoder jointly encodes visual observations, the goal, and embodiment parameters. Action Prediction predicts the action sequence from the fused features. Spatial Perception uses embodiment-aware context and action features to identify high-risk waypoints, while Risk-Aware Correction further refines risky trajectories. }
  \label{fig2}
\end{figure*}

We formulate the navigation task as follows. Given an ego-centric target location $ \mathbf{g} \in \mathbb{R}^2$ and the embodiment parameter $\mathbf{m}$, the agent predicts an $H$-step action sequence based on the image sequence $ I_{t-k:t}$ and the current depth $ D_t$. At each time step, the action is defined as $ \mathbf{a}_t =[v_t, w_t]^\top$, where $v_t$ and $w_t$ denote the linear and angular velocities, respectively.
To this end, the agent learns a policy $\pi_\theta(a_{t:t+H} \mid s_t)$ that maps the current state to a future action sequence, where $a_{t:t+H} = \{\mathbf{a}_i\}_{i=t}^{t+H-1}$.
The state is defined as:
\begin{equation}
s_t = \{I_{t-k:t}, D_{t}, \mathbf{g}, \mathbf{m}\},
\end{equation}
where $s_t$ consists of observations, the relative target location, and the embodiment parameters. During real-world deployment, the relative target $\mathbf{g}$ is updated using onboard odometry, while the pose itself is not directly included in the policy input.

In the context of IL, we construct the overall training dataset $\mathcal{D} = \{(s_t^i, a_{t:t+H}^i)\}_{i=1}^N$. The fundamental objective is to minimize the error between the predicted action sequence and the label:
\begin{equation}
    \mathcal{L}_{\text{BC}}(\theta) = \mathbb{E}_{(s_t, a_{t:t+H}) \sim \mathcal{D}} \left\| \pi_\theta(s_t) - a_{t:t+H} \right\|_2.
\end{equation}
However, relying solely on this standard framework is insufficient. Since the embodiment parameters contained in the dataset consist of limited discrete values, it is difficult for the model to directly learn a generalizable mapping from these geometric features to action control. To address this limitation, we propose a decoupled architecture along with corresponding training objectives. The overall architecture of EA-Nav is illustrated in Figure~\ref{fig2}.


\subsection{Embodiment-Aware Navigation Framework}
\label{sec:architecture}
\textbf{Input Encoder.} 
For a multi-module system, an important goal of the \emph{Input Encoder} is to encode multimodal inputs into a shared context feature that can be reused by downstream modules. In our system, this context feature must remain clean, as it is further consumed by downstream perception and correction modules after \emph{Action Prediction}. If the context feature becomes overly entangled with a predicted high-risk trajectory, it may undermine the effectiveness of subsequent risk perception and trajectory correction.

To address this issue, we adopt a unidirectional cross-attention design. Specifically, we first use pretrained encoders~\cite{simeoni2025dinov3, woo2023convnext} to extract visual and depth features. A dedicated query is then introduced for \emph{Action Prediction} to interact with the context feature through cross-attention, while preventing the query from updating the context feature. In this way, the shared context feature remains focused on observation, goal, and embodiment information, which is crucial for our prediction--perception--correction pipeline.
The multimodal context representation is defined as
\begin{equation}
    z_{\text{context}} = [z_{\text{rgb}}, z_{\text{depth}}, z_g, z_m] \in \mathbb{R}^{N_c \times 384},
\end{equation}
where $N_c = 130$ denotes the total number of context tokens. Unlike prior approaches that compress the inputs into only 8--10 tokens, we retain more tokens to preserve fine-grained spatial information, which is critical for downstream tasks.

\noindent\textbf{Action Prediction.}
This module decodes the action representation $z_{\text{action}}^{*}$ from the \emph{Input Encoder} into an action sequence. To provide an explicit spatial trajectory $z_{\text{traj}}$ for the subsequent \emph{Spatial Perception} module and \emph{Risk-Aware Correction} module, the predicted action sequence is converted into a local waypoint trajectory and further encoded into trajectory features.

\noindent\textbf{Spatial Perception.}
Different from \emph{Action Prediction}, this regression task is more challenging. To accurately estimate the nearest risk distance, the model must capture the relationship between embodiment geometry and the surrounding environment. It must also infer how the scene will evolve under the predicted actions from the current viewpoint. This is the main reason why we introduce an explicit spatial trajectory representation, which helps reduce the difficulty of distance prediction.
In addition, in multi-module architectures, the shared feature is often primarily driven by the main task (e.g., \emph{Action Prediction}), while downstream modules directly reuse this representation for inference. Under such a setting, the shared context feature tends to be biased toward action regression. However, nearest-risk-distance prediction imposes different requirements on the representation, requiring finer-grained and more geometry-aware features. Therefore, we allow the loss of this task to update the shared context feature, which is critical for model convergence and performance. The overall prediction process of the \emph{Spatial Perception} module is formulated as
\begin{equation}
\begin{aligned}
z_{\text{dist}} &= \mathrm{Transformer}\!\left(z^*_{\text{context}}, \mathrm{MLP}(z_{\text{traj}})\right),\\
\hat{\mathbf{d}} &= \mathrm{MLP}(z_{\text{dist}}),
\end{aligned}
\end{equation}
where \(\hat{\mathbf{d}} = [\hat d_1, \hat d_2, \dots, \hat d_H]\) denotes the predicted nearest-obstacle distance for each waypoint. Note that embodiment information has been incorporated into $z^*_{\text{context}}$.

On the other hand, since high-risk samples are relatively scarce in the training data, we design \emph{Spatial Perception} to be decoupled from \emph{Action Prediction}. Specifically, the two modules are organized in a sequential rather than parallel manner, which allows the predicted action sequence to be replaced with augmented high-risk trajectories during training. In contrast, under a parallel design, it would be difficult to directly manipulate and augment such an intermediate representation.
With this design, \emph{Spatial Perception} can learn not only from the widely distributed low-risk trajectories in the dataset, but also from augmented high-risk trajectories that are otherwise scarce. Moreover, since the augmented trajectories are constrained to remain within the current field of view, their associated local scene context remains consistent with the original observation. As a result, the shared context features can be directly reused as a general representation.

\noindent\textbf{Risk-Aware Correction.}
To correct trajectories identified as high-risk by \emph{Spatial Perception}, one possible design is to predict a correction increment action sequence. However, because this module reuses nearly the same context representation as \emph{Action Prediction}, predicting corrections in the same output space would offer limited refinement.
We therefore instead predict a low-dimensional global yaw offset for trajectory correction. This design reduces the prediction difficulty and enables the module to complement, rather than duplicate, the behavior of \emph{Action Prediction}. Concretely, we discretize the correction space within $[-\theta_{\max}, \theta_{\max}]$ into uniformly spaced angle bins, where $\theta_{\max}=45^\circ$ and the bin interval is $5^\circ$. The module predicts the feasibility of each correction bin as
\begin{equation}
\begin{aligned}
z_{\text{corr}} &= \mathrm{Transformer}\!\left(z^*_{\text{context}}, z_{\text{dist}}, \mathrm{MLP}(z_{\text{traj}})\right),\\
\mathbf{p}_{\text{corr}} &= \sigma(\mathrm{MLP}(z_{\text{corr}})).
\end{aligned}
\end{equation}
\begin{figure*}[!t]
  \centering
  \includegraphics[width=\textwidth]{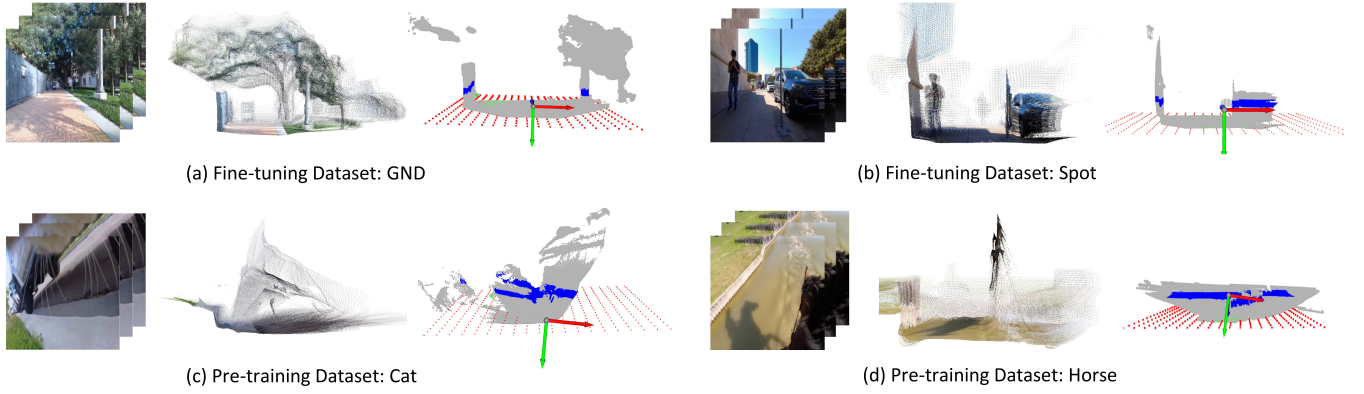}
  \caption{Visualization of environment modeling. The red planar region represents the estimated ground obtained via plane fitting, and the blue regions indicate detected obstacles. The input is an image sequence of length 8.}
  \Description{fig_vis}
  \label{sup:reconstruct}
\end{figure*}
Where $\mathbf{p}_{\text{corr}} \in [0,1]^{\mathcal{B}}$ denotes the predicted feasibility probabilities for all candidate correction bins. During training, we use multi-hot feasible-bin labels as supervision, allowing the model to capture multiple valid recovery directions. During inference, we first obtain the feasible set $\hat{\mathcal{B}}=\{b \in \mathcal{B}\mid \mathbf{p}_{\text{corr}}(b)>\tau\}$, and then select the feasible bin with the smallest deviation from the original high-risk trajectory as the final correction, i.e., $\Delta \hat{\theta} = \arg\min_{b \in \hat{\mathcal{B}}} |b|$.
Compared with single-angle regression, this formulation better captures the one-to-many nature of trajectory correction. In addition, this module follows the same decoupled design as \emph{Spatial Perception}.

\begin{figure}[t]
  \centering
  \includegraphics[width=\linewidth]{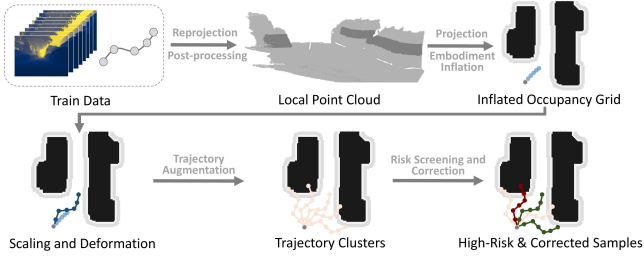}
  \caption{Augmentation pipeline. Light blue: original trajectory; pink: generated trajectory cluster; red: selected collision trajectory; green: corresponding traversable trajectory.}
  \Description{aug}
  \label{fig_aug}
\end{figure}

\subsection{Risk Trajectory Augmentation}
\label{sec:Augmentation}
As illustrated in Figure~\ref{fig_aug}, the pipeline consists of three steps: 1) constructing a local point cloud map, 2) generating an embodiment-conditioned inflated occupancy grid, and 3) synthesizing high-risk trajectories with feasible correction directions.

\noindent\textbf{Point Cloud Map and Inflated Occupancy Grid.}
Given depth observations, trajectories, and camera intrinsics over the next \(H\) timesteps, we fuse them into a robot-centric local point cloud map. Aggregating short-horizon future observations recovers more complete geometry around the trajectory, providing reliable support for high-risk trajectory generation. This process is used only for offline augmentation, not at test time.

We then estimate the ground plane and identify obstacles under embodiment constraints. Valid depth points from the central field of view are used to fit the dominant ground plane with RANSAC. Based on the robot dimensions and maximum traversable height, obstacle regions are inflated to construct an embodiment-aware occupancy grid for subsequent collision checking.

\noindent\textbf{Augmented Trajectory Generation.}
Based on the inflated occupancy grid, we compute a safety coefficient \(s_d\) for the original trajectory:
\begin{equation}
s_d =
\frac{d_{\min}}
{\max\left(\lVert \mathbf{w}_H-\mathbf{w}_0\rVert_2,\epsilon\right)},
\end{equation}
where \(d_{\min}\) is the minimum obstacle clearance and \(\epsilon\) is a small constant. The coefficient \(s_d\) provides a reference scale that moves the trajectory toward the nearest obstacle. Starting from a spline-smoothed trajectory, we uniformly scale all waypoints:
\begin{equation}
\tilde{\mathbf{w}}_i
=
\mathbf{w}_0+\alpha(\mathbf{w}_i-\mathbf{w}_0),
\end{equation}
where \(\alpha\) is sampled around the reference scale \(s_d\). This preserves the trajectory shape while generating candidates with varying obstacle clearances. We further apply rotations from \(-45^\circ\) to \(45^\circ\) at \(5^\circ\) intervals to form a trajectory pool.

Each candidate is checked for collision on the inflated occupancy grid. Colliding candidates are treated as risk trajectories, while collision-free candidates define feasible correction bins. These bins provide multi-hot supervision over the discretized correction space, enabling large-scale supervision from limited fine-tuning data.

\begin{table}[t]
  \centering
  \caption{Comparison of existing navigation datasets.}
  \label{tab:datasets}
  \resizebox{\columnwidth}{!}{%
  \begin{tabular}{lccccc}
    \toprule
    \textbf{Dataset} & \textbf{Embod.} & \textbf{Hours} &
    \textbf{Source} & \textbf{Supervision} & \textbf{Camera} \\
    \midrule
    RECON~\cite{shah2022rapid}
      & 1 & 25 & Real & GT & Fisheye \\
    GoStanford~\cite{hirose2019deep}
      & 1 & 17 & Real & GT & Fisheye \\
    SCAND~\cite{karnan2022socially}
      & 2 & 9 & Real & GT & Pinhole \\
    TartanDrive~\cite{triest2022tartandrive}
      & 1 & 7 & Real & GT & Fisheye \\
    TartanGround~\cite{patel2025tartanground}
      & 3 & $\sim$40 & Synthetic & GT & Pinhole \\
    NavDP~\cite{cai2025navdp}
      & 1 & -- & Synthetic & GT & Pinhole \\
    \midrule
    LeLaN~\cite{hirose2025lelan}
      & 2 & 129 & Real & Pseudo & Fisheye \\
    CityWalker~\cite{liu2025citywalker}
      & 2 & 2000 & Real & Pseudo & Pinhole \\
    FrodoBots-2k~\cite{hirose2025learning}
      & 2 & 2000 & Real & Pseudo & Fisheye \\
    OmniVLA~\cite{hirose2025omnivla}
      & 3 & 9500 & Real & Pseudo & Fish\&Pin \\
    \rowcolor{cyan!10}
    \textbf{Ours}
      & 8 & 1000 & Real & Pseudo & Pinhole \\
    \bottomrule
  \end{tabular}%
  }
\end{table}

\subsection{Embodiment-Aware Training Framework}
\label{sec:training}

\noindent\textbf{Training Data Construction.}
For pretraining, we collect heterogeneous first-person videos from the Internet and use Depth-Anything-3~\cite{lin2025depth} to construct samples with consistent observation and trajectory scales. Exact metric alignment is unnecessary, as navigation training only requires consistency between the estimated trajectory and observation scales. The embodiment geometry of each category is manually annotated based on its category attributes. To reduce noise, we filter samples by trajectory smoothness and discard those with unreliable labels. As shown in Table~\ref{tab:datasets}, the resulting dataset contains approximately 1,000 hours of real-world first-person videos across eight embodiment categories, providing both large-scale data and diverse embodiments.

For fine-tuning, we construct physically consistent depth observations using Depth-Anything-3~\cite{lin2025depth}, conditioned on metric-scale trajectories and camera intrinsics from real-world data. The embodiment geometry is obtained from the physical parameters of each platform. As illustrated in Figure~\ref{sup:reconstruct}, short image sequences are fused into local point clouds, from which the ground plane and obstacle regions are extracted to construct geometric supervision.

\noindent\textbf{Pretraining Stage.}
We train the base policy on the constructed pretraining dataset. Given an image sequence, current depth observation, target location, and embodiment geometry, the model predicts an action sequence. Since Internet videos contain substantial noise and cannot provide reliable geometric supervision, we optimize only the action prediction branch using behavior cloning.

\noindent\textbf{Fine-tuning Stage.}
During fine-tuning, we train the model on real-world data augmented by the proposed \emph{Risk Trajectory Augmentation}. The action prediction branch is supervised by real actions using behavior cloning, while \emph{Spatial Perception} and \emph{Risk-Aware Correction} are supervised by occupancy-grid targets and feasible correction bins, respectively. The three objectives are jointly optimized using a weighted sum.

\section{Experiment}
\subsection{Experimental Setup}
\noindent\textbf{Datasets.}
Pre-training is conducted on our proposed heterogeneous dataset. To improve coverage of indoor environments, we additionally incorporate a mini version of HM3D~\cite{cai2025navdp} as supplementary data. During fine-tuning, GND~\cite{liang2025gnd}, SCAND-Spot~\cite{karnan2022socially}, and SiT~\cite{bae2023sit} are used for supervision. For evaluation, we report quantitative results on i2Nav~\cite{tang2025i2nav} and further validate the effectiveness of our method through real-world robot experiments.

\noindent\textbf{Evaluation Metrics.}
We evaluate our method from three perspectives: Spatial Perception, Risk-Aware Correction, and Embodiment-Aware Navigation. For spatial perception, we use TPR (True Positive Rate), FAR (False Alarm Rate), and MAE (Mean Absolute Error) to evaluate the model's ability to perceive risk. An action sequence is regarded as high-risk if the predicted minimum obstacle distance over its waypoints is smaller than the safety threshold $\delta_s = 0.5\,\mathrm{m}$.

For the Risk-Aware Correction, we use CSR (Correction Success Rate) and IoU to evaluate correction effectiveness and the quality of feasible-bin prediction. The $IoU_{bin}$ is defined as a micro IoU over all high-risk samples.
For the overall navigation performance, we report SR (Success Rate) and CR (Collision Rate).

\subsection{Embodiment Modality in Pre-training}
\begin{figure}[t]
  \centering
  \includegraphics[width=0.97\linewidth]{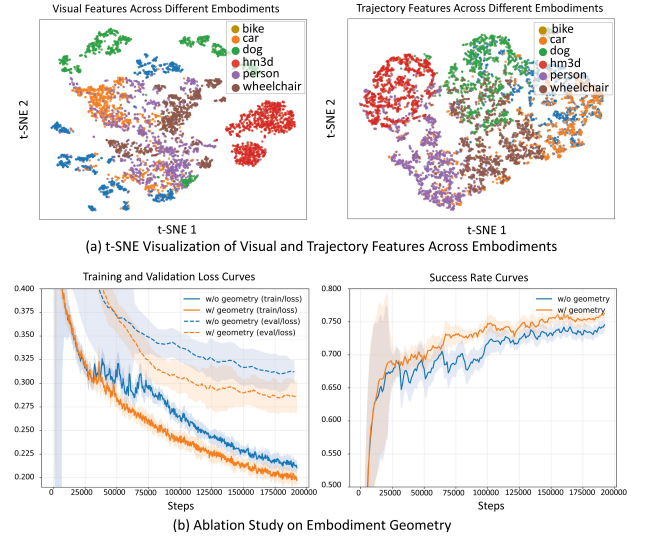}
  \caption{Analysis of embodiment geometry in pre-training.}
  \Description{fig_vis}
  \label{fig_vis}
\end{figure}

To investigate the role of introducing embodiment geometry during pre-training, we conduct both qualitative and quantitative analyses.
First, we examine whether navigation data from different embodiments exhibit inherent behavioral ambiguity under similar visual observations. To this end, we sample a subset from the heterogeneous pre-training dataset constructed in this work, where each embodiment category contains 1,000 image--motion clips. We use a frozen pretrained DINOv3 encoder to represent the image sequences and visualize the resulting visual features with t-SNE~\cite{van2008visualizing}. Meanwhile, we further construct a set of motion attributes, including straightness, speed standard deviation, stop-and-go ratio, mean acceleration, and mean yaw change, and similarly visualize them using t-SNE. As shown in Figure~\ref{fig_vis}(a), although some embodiments are close to each other in the visual observation space, their motion attribute distributions still differ significantly. For example, wheelchair and person samples may overlap in the visual feature space, while exhibiting distinct patterns in the motion attribute space. This suggests that, without embodiment geometry as a conditioning signal, the model faces ambiguity in mapping observations to actions, and is therefore more likely to learn an averaged fit over multiple behavioral modes.

Based on this observation, we further conduct an ablation study on embodiment geometry during pre-training. Specifically, we remove the embodiment geometry input and pre-train the model using only visual observations and goal locations, while keeping all other training settings unchanged. We then compare this variant with the full model. As shown in Figure~\ref{fig_vis}(b), introducing embodiment geometry consistently improves the training loss, validation loss, and test-set SR, indicating that explicit embodiment modeling benefits both policy learning and cross-embodiment generalization.

Notably, although introducing embodiment geometry yields consistent improvements, removing this modality does not cause a dramatic performance drop. We conjecture that image sequences can still implicitly encode part of the embodiment information. For example, pixel changes across adjacent frames may reflect the motion speed range, while viewpoint height may indirectly correlate with the geometric characteristics of different embodiments. Nevertheless, the experimental results show that explicitly modeling embodiment geometry remains beneficial for multi-embodiment visual navigation, particularly for models that do not take temporal image sequences as input.

\subsection{Distance Prediction Performance}
To evaluate spatial perception, we conduct experiments on the i2Nav~\cite{tang2025i2nav} dataset using five agent sizes with embodiment radii of 0.5\,m, 1.0\,m, 1.5\,m, 2.0\,m, and 2.5\,m. Equal length and width are used only to standardize the evaluation, while our method supports anisotropic embodiments. The Body-L setting (2.5\,m) lies outside the training range and evaluates generalization.

\begin{figure}[t]
  \centering
  \includegraphics[width=\linewidth]{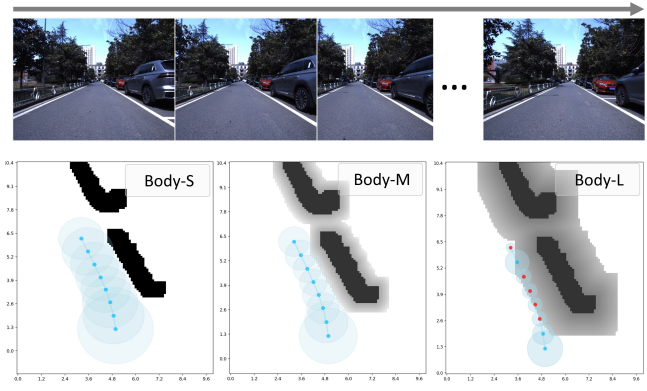}
  \caption{Qualitative analysis of spatial perception performance. Blue points indicate collision-free waypoints, while red points indicate collision waypoints.}
  \Description{fig_vis}
  \label{distance}
\end{figure}

As shown in Figure~\ref{distance}, different embodiment sizes produce different minimum obstacle distances along the same trajectory, indicating that the model captures embodiment-conditioned spatial risk. Table~\ref{tab:distance_perception} further shows strong performance across all five sizes, including the out-of-range Body-L setting. Compared with training without augmentation, risk augmentation improves the recognition of high-risk and collision samples by about $5\times$ with almost no additional false alarms. These results demonstrate the effectiveness of the augmentation strategy and the robustness of the model across embodiment sizes.

\begin{table}[t]
\centering
\caption{Spatial Perception Performance}
\label{tab:distance_perception}
\resizebox{\columnwidth}{!}{%
\begin{threeparttable}
\begin{tabular}{ccccccccc}
\toprule
\multirow{2}{*}[-1.0ex]{\textbf{\makecell{Size}}}
 & \multicolumn{4}{c}{\textbf{w/ Aug}} & \multicolumn{4}{c}{\textbf{w/o Aug}} \\
\cmidrule(lr){2-5} \cmidrule(lr){6-9}
 & TPR$\uparrow$ & TPR$^{*}$$\uparrow$ & FAR$\downarrow$ & MAE$\downarrow$ & TPR$\uparrow$ & TPR$^{*}$$\uparrow$ & FAR$\downarrow$ & MAE$\downarrow$ \\
\midrule
Body-S\phantom{$^+$} & \textbf{84.9} & \textbf{89.6} & \textbf{0.00} & \textbf{0.471} & 12.9 & 13.4 & \textbf{0.00} & 0.664 \\
Body-S$^+$           & \textbf{79.7} & \textbf{92.3} & 0.80 & \textbf{0.511} & 8.8 & 10.0 & \textbf{0.00} & 0.715 \\
Body-M\phantom{$^+$} & \textbf{79.4} & \textbf{95.1} & 1.00 & \textbf{0.478} & 11.1 & 13.5 & \textbf{0.00} & 0.711 \\
Body-M$^+$           & \textbf{70.6} & \textbf{85.0} & 2.50 & \textbf{0.532} & 21.6 & 28.2 & \textbf{1.06} & 0.712 \\
Body-L\phantom{$^+$} & \textbf{73.8} & \textbf{83.5} & 5.30 & \textbf{0.582} & 29.2 & 35.8 & \textbf{0.00} & 0.799 \\
\bottomrule
\end{tabular}
\begin{tablenotes}[flushleft]
\footnotesize
\item[\textbullet] TPR is computed on high-risk samples (distance $<$ 0.5 m). 
\item[\textbullet] TPR$^{*}$ is computed on collision samples only (distance $<$ 0 m), at the same decision threshold.
\item[\textbullet] The units of TPR and FAR are \%, and the unit of MAE is m.
\end{tablenotes}
\end{threeparttable}%
}
\end{table}

\subsection{Trajectory Correction Performance}
To evaluate the risk-aware correction of our model, we conduct quantitative experiments on the i2Nav~\cite{tang2025i2nav} dataset. Since collision samples are scarce in real-world data, we use the risk augmentation module to generate collision trajectories and their corresponding correction targets. Consistent with Sec.~4.3, we evaluate the model under five embodiment sizes.

As shown in Table~\ref{tab:correction}, our method achieves strong embodiment-aware trajectory correction performance across different embodiment settings, showing that the model can leverage both observation cues and embodiment geometry for risk-aware refinement. We further compare our method with a random correction baseline that uniformly samples a deflection angle from $[-45^\circ, 45^\circ]$. Under small-embodiment settings (e.g., Body-S), the random strategy can achieve higher CSR because the feasible correction region is relatively large, allowing many random angles to succeed. However, this comes at the cost of much larger correction angles and limited controllability. Our method is less advantageous in such cases because obstacles are relatively farther away for smaller embodiments along the same trajectory. As a result, correction becomes more dependent on far-range correction ability, where the current model remains limited. As the embodiment size increases and the feasible region shrinks, the random strategy degrades rapidly. In contrast, our method remains more stable while keeping the correction angle within a relatively small range, demonstrating better controllability and embodiment-aware refinement capability.

In addition, we introduce $IoU_{bin}$ to evaluate the model's understanding of feasible deflection angles. As shown in Table~\ref{tab:correction}, our method achieves high $IoU_{bin}$ across different embodiment settings, indicating that it can identify feasible correction regions in angle space. Notably, $IoU_{bin}$ is not fully consistent with CSR. Under small-embodiment settings, the model can obtain relatively high $IoU_{bin}$ while CSR remains lower. This is because obstacles are farther away in such cases, resulting in wider feasible intervals. As a result, assigning high feasibility scores to multiple deflection angles can already produce high region overlap. However, a high $IoU_{bin}$ does not necessarily mean that the model can accurately identify the truly effective correction angles. In contrast, in near-range high-risk scenarios, although the feasible interval is harder to recover completely, the model can often identify the most effective correction angles more accurately, leading to higher correction success.

\begin{table}[t]
\centering
\caption{Embodiment-Aware Trajectory Refinement}
\label{tab:correction}
\resizebox{\columnwidth}{!}{%
\begin{threeparttable}
\begin{tabular}{ccccccc}
\toprule
\multirow{2}{*}[-1.0ex]{\textbf{\makecell{Size}}}
 & \multicolumn{3}{c}{\textbf{Ours}} & \multicolumn{3}{c}{\textbf{Random}} \\
\cmidrule(lr){2-4} \cmidrule(lr){5-7}
& CSR(\%)$\uparrow$& MDA($^\circ$)$\downarrow$ & $\text{IoU}_{\text{bin}}$(\%)$\uparrow$  & CSR(\%)$\uparrow$ & MDA($^\circ$)$\downarrow$ & $\text{IoU}_{\text{bin}}$(\%)$\uparrow$ \\
\midrule
Body-S\phantom{$^+$} & 57.00 & \textbf{12.30} & 70.97 & \textbf{66.21} & 22.01 & -- \\
Body-S$^+$ & \textbf{55.52} & \textbf{15.41} & 60.87 & 49.85 & 25.28 & -- \\
Body-M\phantom{$^+$} & \textbf{57.19} & \textbf{13.00} & 59.30 & 42.36 & 23.20 & -- \\
Body-M$^+$ & \textbf{58.61} & \textbf{13.42} & 59.34 & 41.11 & 24.58 & -- \\
Body-L\phantom{$^+$} & \textbf{65.97} & \textbf{18.29} & 53.08 & 39.83 & 22.59 & -- \\
\bottomrule
\end{tabular}%
\begin{tablenotes}[flushleft]
\footnotesize
\item[\textbullet] For clarity, the $IoU_{bin}$ is computed over the central 19 bins.
\item[\textbullet] “--” denotes that $IoU_{bin}$ is undefined for a random policy.
\end{tablenotes}
\end{threeparttable}
}
\end{table}

\begin{table}[t]
\centering
\caption{Navigation performance in simulation environments}
\label{tab:simulation}
\small
\setlength{\tabcolsep}{3.5pt} 
\renewcommand{\arraystretch}{0.9}
\resizebox{0.92\columnwidth}{!}{%
\begin{threeparttable}
\begin{tabular}{lcccccc}
\toprule
\multicolumn{1}{c}{%
  \multirow{2}{*}{\textbf{Methods}}%
}
& \multicolumn{2}{c}{\textbf{Scene1}}
& \multicolumn{2}{c}{\textbf{Scene2}}
& \multicolumn{2}{c}{\textbf{Scene3}} \\

\cmidrule(lr){2-3}
\cmidrule(lr){4-5}
\cmidrule(lr){6-7}
& SR$\uparrow$ & SPL$\uparrow$
& SR$\uparrow$ & SPL$\uparrow$
& SR$\uparrow$ & SPL$\uparrow$ \\
\midrule
iPlanner~\cite{yang2023iplanner}
& 0.66 & 0.54
& 0.50 & 0.39
& 0.30 & 0.18 \\
NavDP~\cite{cai2025navdp}
& \textbf{0.73} & \textbf{0.63}
& \underline{0.60} & \textbf{0.52}
& \underline{0.56} & \underline{0.41} \\
NoMaD$^*$~\cite{sridhar2024nomad}
& 0.55 & 0.39
& 0.44 & 0.31
& 0.36 & 0.22 \\
ExAug$^*$~\cite{hirose2023exaug}
& 0.61 & 0.42
& 0.46 & 0.33
& 0.38 & 0.24 \\
\rowcolor{cyan!10}
\textbf{Ours} (w/o corr)
& 0.62 & 0.51
& 0.45 & 0.35
& 0.43 & 0.31 \\
\rowcolor{cyan!10}
\textbf{Ours} (w/ corr)
& \underline{0.70} & \underline{0.59}
& \textbf{0.62} & \underline{0.51}
& \textbf{0.60} & \textbf{0.49} \\
\bottomrule
\end{tabular}
\end{threeparttable}%
}
\end{table}

\subsection{Embodiment-Aware Navigation Evaluation}
We evaluate the proposed embodiment-aware navigation system in simulation on InternUtopia~\cite{wang2024grutopia} and in the real world using TurtleBot and Unitree Go2. Different embodiment sizes are configured for each robot, as shown in Figure~\ref{fig:robot}.

\begin{figure*}[!t] 
    \centering
    \includegraphics[width=\textwidth]{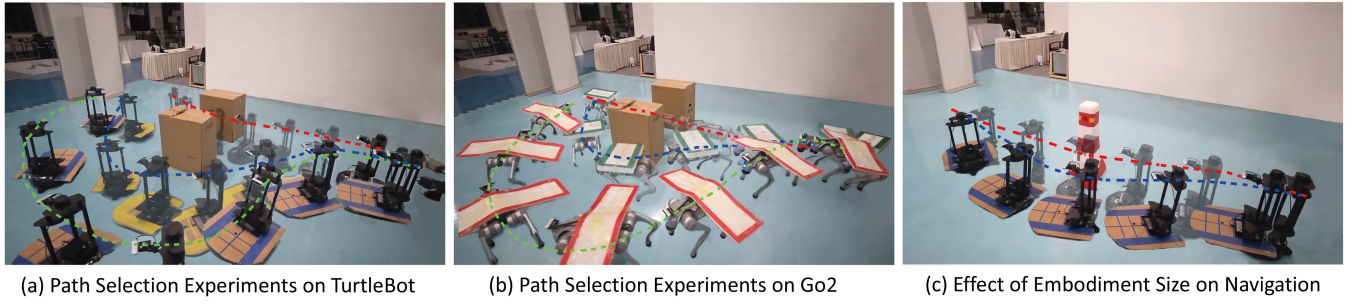}
    \caption{Embodiment-Aware Navigation Strategies Across Different Robots. Smaller robots tend to navigate through narrow spaces, while larger robots prefer more open paths.}
    \label{fig:vis_bot}
\end{figure*}

\begin{figure}[t]
    \centering
    \includegraphics[width=\linewidth, height=0.4\textheight, keepaspectratio]{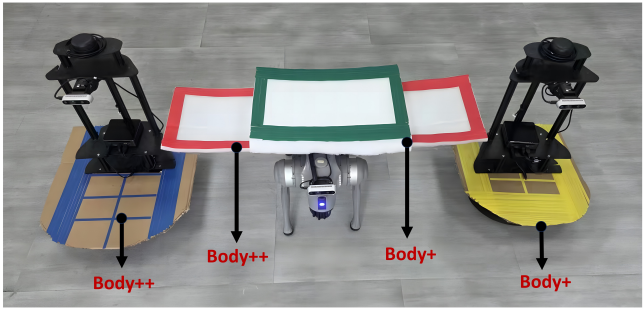}
    \caption{Real-world experimental setup. Body denotes the original agent size, while Body+ and Body++ represent enlarged embodiments.}
    \label{fig:robot}
\end{figure}

\noindent\textbf{Evaluation in Simulation Environments.}
We use the easy, hard, and indoor assets from NavDP~\cite{cai2025navdp} and InternUtopia~\cite{wang2024grutopia}, denoted as Scene1, Scene2, and Scene3 in Table~\ref{tab:simulation}. Navigation becomes increasingly difficult as obstacle density increases.
Our base model, Ours(w/o corr), underperforms NavDP, mainly because our training data are collected primarily from real-world environments and are smaller and less diverse than the large-scale synthetic data used by NavDP. This results in a noticeable sim-to-real distribution gap.
Adding embodiment-aware Spatial Perception and Risk-Aware Correction improves the average performance by approximately 31\%, with larger gains in the more cluttered Scene2 and Scene3. These results show that embodiment conditions and task-specific inductive biases can effectively compensate for limited training data, particularly in complex environments. Comparing Ours (w/o corr) and Ours (w/ corr) in Table~\ref{tab:simulation} further shows that the perception and correction modules provide larger gains in cluttered scenes.

\begin{table}[t]
\centering
\caption{Real-world Navigation Performance Evaluation}
\label{tab:real}
\renewcommand{\arraystretch}{0.92}
\resizebox{0.92\columnwidth}{!}{%
\begin{threeparttable}
\begin{tabular}{l|lcccccc}
\toprule
\multicolumn{2}{c}{\multirow{2}{*}{\textbf{Methods}}}
& \multicolumn{2}{c}{\textbf{Body}}
& \multicolumn{2}{c}{\textbf{Body+}}
& \multicolumn{2}{c}{\textbf{Body++}} \\
\cmidrule(lr){3-4}
\cmidrule(lr){5-6}
\cmidrule(lr){7-8}
\multicolumn{2}{c}{}
& SR$\uparrow$ & CR$\downarrow$
& SR$\uparrow$ & CR$\downarrow$
& SR$\uparrow$ & CR$\downarrow$ \\
\midrule

& iPlanner~\cite{yang2023iplanner}
& 0.20 & 1.00
& 0.20 & 1.00
& 0.00 & 1.00 \\

& NavDP~\cite{cai2025navdp}
& \underline{0.60} & \underline{0.60}
& \underline{0.40} & \underline{0.80}
& \underline{0.00} & \underline{1.00} \\

& NoMaD*~\cite{sridhar2024nomad}
& 0.20 & 0.20
& 0.00 & 0.40
& 0.00 & 1.00 \\

& ExAug*~\cite{hirose2023exaug}
& 0.20 & 0.20
& 0.20 & 0.40
& 0.00 & 1.00 \\

\rowcolor{cyan!10}
\cellcolor{white}
\multirow{-5}{*}{\rotatebox{90}{Unitree Go2}}
& \textbf{Ours}
& \textbf{0.80} & \textbf{0.20}
& \textbf{0.60} & \textbf{0.20}
& \textbf{0.60} & \textbf{0.00} \\

\midrule

& iPlanner~\cite{yang2023iplanner}
& 0.60 & 0.40
& 0.40 & 0.80
& 0.00 & 1.00 \\

& NavDP~\cite{cai2025navdp}
& \textbf{0.80} & \textbf{0.00}
& \underline{0.60} & \underline{0.60}
& 0.00 & 1.00 \\

& NoMaD*~\cite{sridhar2024nomad}
& 0.20 & 0.20
& 0.00 & 0.60
& 0.00 & 0.60 \\

& ExAug*~\cite{hirose2023exaug}
& 0.20 & 0.20
& 0.20 & 0.60
& 0.00 & \underline{0.20} \\

\rowcolor{cyan!10}
\cellcolor{white}
\multirow{-5}{*}{\rotatebox{90}{TurtleBot}}
& \textbf{Ours}
& \underline{0.60} & \underline{0.20}
& \textbf{0.60} & \textbf{0.40}
& \textbf{0.40} & \textbf{0.20} \\

\bottomrule
\end{tabular}

\begin{tablenotes}[flushleft]
\footnotesize
\item[\textbullet] \textbf{Bold} indicates the best result, and
\underline{underline} indicates the second-best result.
\item[\textbullet] $^*$ denotes image-goal navigation methods.
\end{tablenotes}
\end{threeparttable}%
}
\end{table}

\noindent\textbf{Real-World Experimental Evaluation.}
We first examine whether the model selects paths according to embodiment geometry. As shown in Figure~\ref{fig:vis_bot}(a) and (b), the smaller embodiment chooses a narrow path, whereas the larger one selects a wider route. In Figure~\ref{fig:vis_bot}(c), the smaller embodiment passes directly between nearby obstacles, while the larger one makes a detour. These examples demonstrate that the model can adapt its navigation behavior to embodiment size.
We further conduct quantitative experiments in real-world environments, with five trials for each method. The evaluation combines the scenarios in Figure~\ref{fig:vis_bot}(a)--(c), with additional details provided in the supplementary material. As shown in Table~\ref{tab:real}, our method maintains a high SR and a low CR across different settings, whereas competing methods degrade noticeably.

Point-goal methods generally exhibit higher collision rates because of their aggressive goal-directed behavior. Image-goal methods are safer but degrade substantially in low-texture environments and may lose track of the target. These results highlight the importance of embodiment geometry for point-goal navigation, especially in practical real-world deployment.

\vspace{0.4\baselineskip}
\section{Conclusion and Limitation}

\noindent\textbf{Conclusion.} In this work, we propose an IL-based embodiment-aware visual navigation framework that incorporates embodiment geometry into the navigation model. In the pretraining stage, we construct a heterogeneous cross-embodiment dataset from Internet videos and inject embodiment geometry into the model to learn robust embodiment-related priors. In the fine-tuning stage, we introduce an augmentation strategy to address the scarcity of high-risk samples, while jointly enhancing embodiment-aware spatial perception and risk-aware correction capabilities.

\noindent\textbf{Limitation.} Our current method does not comprehensively model a wider range of embodiment-related factors, such as turning radius and motion constraints, which limits real-world deployment across embodiments with different motion patterns. Future work will focus on incorporating more comprehensive embodiment attributes and enabling more fine-grained embodiment awareness.

\vspace{0.4\baselineskip}
\begin{acks}
This work was supported in part by the ``Pioneer and Leading Goose'' R\&D Program of Zhejiang Province, China, under Grant 2023C01067.
\end{acks}

\clearpage
\bibliographystyle{ACM-Reference-Format}
\bibliography{main}

\appendix
\clearpage
\section{Appendix}
\label{sup:sec:supplementary}

\providecommand{\rowcolor}[1]{}


\subsection{Overview}

This appendix is organized as follows.
Section~\ref{sup:sec:Implementation} provides the implementation details.
Section~\ref{sup:sec:Dataset} describes the construction of the proposed
cross-embodiment dataset.
Section~\ref{sup:sec:Environment} introduces the simulation and real-world
evaluation environments.
Section~\ref{sup:sec:Design} presents the model architecture,
hyperparameter settings, and ablation study.
Finally, Section~\ref{sup:sec:Geometric} discusses the geometric
representation used during fine-tuning.


\subsection{Implementation Details}
\label{sup:sec:Implementation}

For pre-training, we use Adam with a batch size of 64, a weight decay of
$1\times10^{-4}$, and a cosine learning rate schedule from $1\times10^{-4}$
to $1\times10^{-5}$ over 500k steps.
The RGB encoder DINOv3-S~\cite{simeoni2025dinov3} is frozen, while the
depth encoder ConvNeXt-T~\cite{woo2023convnext} is initialized from
pretrained weights.
Both RGB and depth inputs are resized to $224\times224$, with the history
length and prediction horizon set to $k=4$ and $H=8$, respectively.
Pre-training is conducted on four NVIDIA H20 GPUs.
For fine-tuning, we initialize the model from the pretrained checkpoint
and use a batch size of 32 with the same weight decay and learning rate
schedule.
The visual encoders remain frozen, and all experiments are conducted on
a single NVIDIA RTX 4090 GPU.


\subsection{Cross-Embodiment Dataset Construction}
\label{sup:sec:Dataset}

To construct a multi-embodiment navigation dataset, we collect
first-person videos from YouTube, covering animals, humans, and vehicles
across eight embodiment categories.
Unlike prior work~\cite{liu2025citywalker}, we argue that navigation
pre-training data only require scale consistency among trajectories,
observations, and camera intrinsics, rather than strict alignment with
the true metric scale.

\begin{figure}[t]
  \centering
  \includegraphics[width=\linewidth]{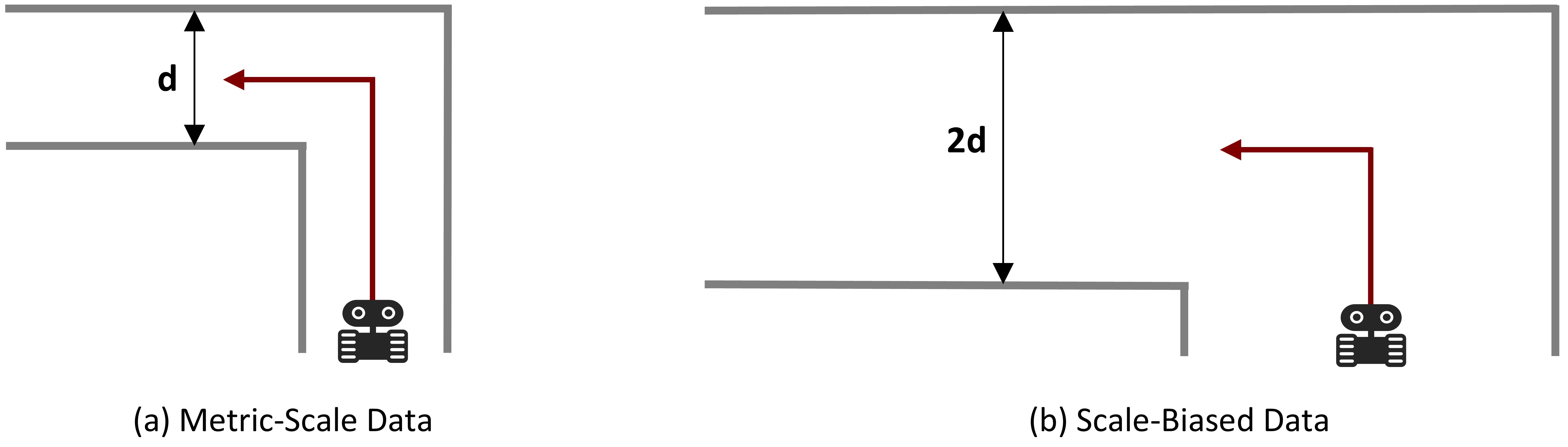}
  \caption{Effect of scale deviation on navigation training samples.
  $d$ denotes the true scale, while $2d$ denotes a twofold scale
  deviation.}
  \Description{Illustration of navigation samples under true and
  twofold-deviated scales.}
  \label{sup:fig1}
\end{figure}

As illustrated in Figure~\ref{sup:fig1}, even when the annotated scale
deviates from the real-world scale by a factor of two, the resulting
samples still induce similar turning behaviors and mainly differ in
their relative distances to obstacles.
Figure~\ref{sup:fig1}(b) can therefore be regarded as an augmented sample
derived from Figure~\ref{sup:fig1}(a), while preserving reasonable
observations and trajectories.
Based on this observation, we use
Depth-Anything-3~\cite{lin2025depth} to estimate depth, trajectories,
and camera intrinsics for the collected video segments.
We manually curate videos whose motion patterns are compatible with
robot navigation and apply a trajectory smoothness check to remove
unreliable samples, including videos with abrupt viewpoint changes or
severe occlusions.

We also evaluate several automatic annotation methods, including
geometry-based approaches such as MegaSaM~\cite{li2025megasam} and
Vipe~\cite{huang2025vipe}, as well as data-driven methods such as
VGGT~\cite{wang2025vggt} and
Depth-Anything-3~\cite{lin2025depth}.
Considering both robustness and annotation accuracy, we adopt
Depth-Anything-3 as the core model for automatic annotation.

\subsection{Evaluation Environment Setup}
\label{sup:sec:Environment}

\begin{figure}[t]
  \centering
  \includegraphics[width=0.7\linewidth]{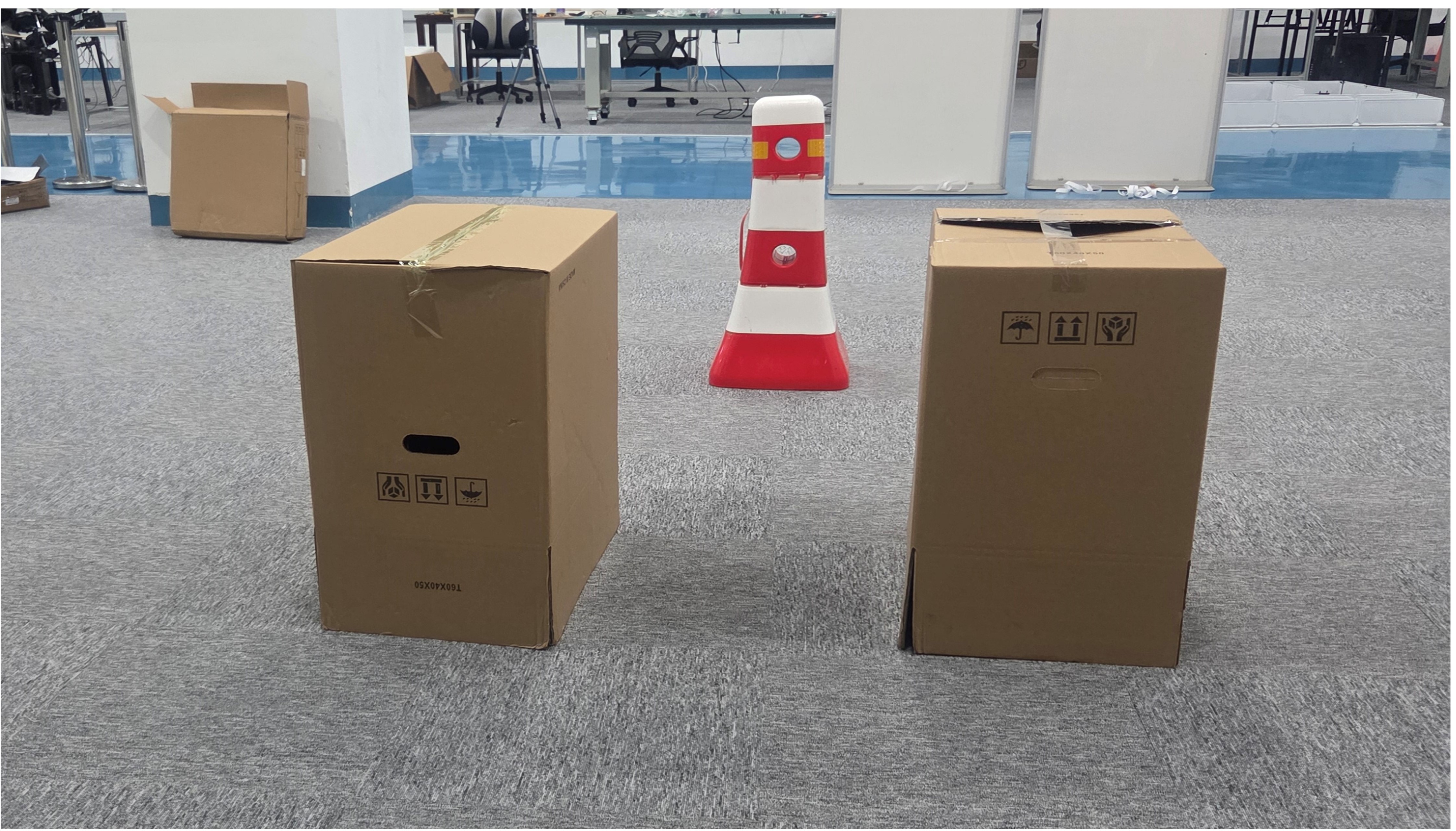}
  \caption{Layout of the real-world evaluation environment.
  Cardboard boxes are treated as fixed obstacles, while road barriers
  are regarded as movable obstacles.}
  \Description{Real-world navigation environment containing cardboard
  boxes and a movable road barrier.}
  \label{sup:fig4}
\end{figure}

\begin{figure}[t]
  \centering
  \includegraphics[width=\linewidth]{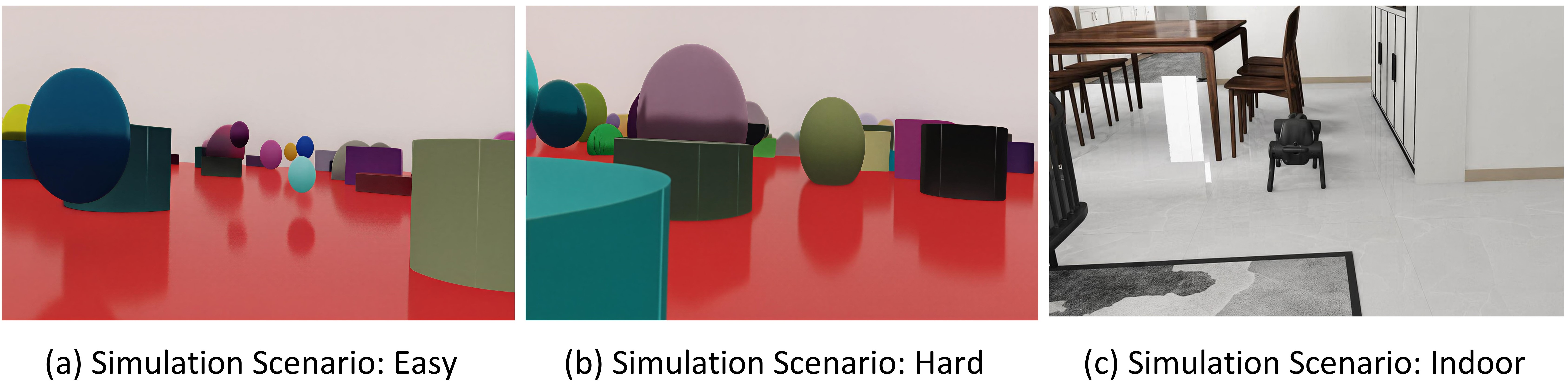}
  \caption{Simulation assets used for evaluation, including scenes with
  different obstacle layouts and difficulty levels.}
  \Description{Easy, hard, and indoor simulation environments used for
  navigation evaluation.}
  \label{sup:fig3}
\end{figure}

\noindent\textbf{Simulation Environments.}
We use three types of simulation assets, as shown in
Figure~\ref{sup:fig3}.
Figures~\ref{sup:fig3}(a) and~\ref{sup:fig3}(b) correspond to the
\textit{Easy} and \textit{Hard} assets provided by
NavDP~\cite{cai2025navdp}, respectively, where the target position is
randomly assigned in each episode.
We additionally evaluate the model in the indoor simulation environment
shown in Figure~\ref{sup:fig3}(c).
These environments contain different obstacle layouts and difficulty
levels, enabling a comprehensive evaluation of navigation performance.

\noindent\textbf{Real-World Environment.}
For real-world evaluation, we construct the environment shown in
Figure~\ref{sup:fig4}.
Cardboard boxes are treated as fixed obstacles, and a frontal collision
with them is regarded as a navigation failure.
The red road barrier is treated as a movable obstacle that the agent is
allowed to contact during navigation.
The target is placed 4 meters ahead of the agent.
This setup evaluates the navigation success and collision behavior of
different models under real-world conditions.


\subsection{Model Architecture and Ablation Details}
\label{sup:sec:Design}

\noindent\textbf{Model Configuration.}
Table~\ref{tab:network_architecture} summarizes the model architecture and
hyperparameter settings.
The model adopts a unified token-based representation, where RGB, depth,
goal, and embodiment geometry are projected into a shared
384-dimensional feature space.
A Transformer-based input encoder fuses the multimodal context tokens,
followed by decoupled modules for Action Prediction, Spatial Perception,
and Risk-Aware Correction.

\clearpage


\begin{table}[H]
\centering
\caption{Ablation study on model components.
Sub-Grad. denotes updating context tokens using sub-module losses,
while Uni-Mask denotes applying a unidirectional mask.}
\label{sup:tab:ablation}

\resizebox{\linewidth}{!}{%
\begin{tabular}{ccccc}
\toprule
\multirow{2}{*}{\textbf{Sub-Grad.}}
&
\multirow{2}{*}{\textbf{Uni-Mask}}
&
\multicolumn{3}{c}{\textbf{Test}} \\
\cmidrule(lr){3-5}
&
&
Pose Loss
&
Percept. Loss
&
Corr. Loss \\
\midrule
            & \checkmark & 0.055          & 0.020          & 0.195          \\
\checkmark  & \checkmark & 0.051          & \textbf{0.012} & \textbf{0.162} \\
\checkmark  &            & \textbf{0.050} & 0.015          & 0.174          \\
\bottomrule
\end{tabular}%
}
\end{table}

The history length is set to $k=4$, and the prediction horizon is set to
$H=8$ to balance temporal modeling capacity and computational
efficiency.
For risk-aware correction, the angular range $[-45^\circ,45^\circ]$ is
discretized at $5^\circ$ intervals, and threshold $\tau$ is used to
select feasible corrections.

\noindent\textbf{Ablation on Sub-Grad and Uni-Mask.}
Table~\ref{sup:tab:ablation} evaluates the effects of Sub-Grad and
Uni-Mask.
Allowing sub-module gradients to update the context tokens reduces the
mismatch between the coarse-grained features learned for action
prediction and the features required by downstream tasks.
Applying a unidirectional mask prevents downstream tokens from
interfering with the shared context representation, further improving
the spatial perception and risk-aware correction modules.


\subsection{Geometric Representation of the Environment}
\label{sup:sec:Geometric}

Accurate environmental geometry is important for embodiment-aware
navigation.
However, we introduce explicit geometric modeling only during
fine-tuning rather than pre-training.
The first-person videos used for pre-training often contain rapid,
irregular, or non-rigid motion.
For example, the movement of an animal's head may be inconsistent with
the overall displacement of its body.
Such motion violates the rigid-scene and stable-camera assumptions
required by camera-based geometric reconstruction, resulting in noisy
depth and scene geometry.

These reconstruction errors impair ground-region identification and
camera-height estimation, making it difficult to reliably separate
obstacles from traversable space.
Consequently, geometric annotations generated from unconstrained
Internet videos may provide misleading supervision during pre-training.
In contrast, the fine-tuning data are collected by robots in relatively
structured environments, where camera viewpoints and motion patterns
are more stable.
The annotation model can therefore reconstruct the environment more
reliably and provide more accurate geometric supervision for spatial
perception and risk-aware navigation.

\newpage


\begin{table}[H]
\centering
\caption{Network architecture and hyperparameter settings.}
\label{tab:network_architecture}

\begin{tabularx}{\linewidth}{@{}lX@{}}
\toprule
\textbf{Item} & \textbf{Setting} \\
\midrule

\multicolumn{2}{l}{\textit{Input Encoder}} \\
\midrule

RGB encoder
& DINOv3-S~\cite{simeoni2025dinov3} \\

Depth encoder
& ConvNeXt-T~\cite{woo2023convnext} \\

Input resolution
& $224 \times 224$ \\

RGB tokens $z_{\mathrm{rgb}}$
& $64 \times 384$ \\

Depth tokens $z_{\mathrm{depth}}$
& $64 \times 384$ \\

Goal token $z_g$
& $1 \times 384$ \\

Geometry token $z_m$
& $1 \times 384$ \\

Action query tokens $z_{\mathrm{action}}$
& $8 \times 384$ \\

Context tokens $N_c$
& $130 \times 384$ \\

History length $k$
& $4$ \\

Prediction horizon $H$
& $8$ \\

Transformer encoder
& 6 layers, 384-dim \\

\midrule
\multicolumn{2}{l}{\textit{Action Prediction}} \\
\midrule

Prediction network
& 2-layer MLP, $8 \times 384 \rightarrow 8 \times 3$ \\

\midrule
\multicolumn{2}{l}{\textit{Spatial Prediction}} \\
\midrule

Trajectory encoder
& 2-layer MLP, $8 \times 2 \rightarrow 8 \times 384$ \\

Transformer encoder
& 3 layers, 384-dim \\

Distance tokens $z_{\mathrm{dist}}$
& $8 \times 384$ \\

Prediction network
& 3-layer MLP, $8 \times (384+2+4) \rightarrow 8 \times 1$ \\

Risk threshold $\delta_s$
& $0.5\,\mathrm{m}$ \\

\midrule
\multicolumn{2}{l}{\textit{Risk-Aware Correction}} \\
\midrule

Trajectory encoder
& 2-layer MLP, $8 \times 2 \rightarrow 8 \times 384$ \\

Transformer encoder
& 3 layers, 384-dim \\

Correction tokens $z_{\mathrm{corr}}$
& $8 \times 384$ \\

Distance summary token
& Attention pooling, $8 \times 384 \rightarrow 1 \times 384$ \\

Token aggregation
& Attention pooling, $9 \times 384 \rightarrow 1 \times 384$ \\

Prediction network
& 3-layer MLP, $1 \times 384 \rightarrow 1 \times 19$ \\

Correction angle range
& $[-45^\circ,\,45^\circ]$ \\

Correction bin interval
& $5^\circ$ \\

Number of correction bins
& 19 \\

Decoding rule
& Argmax over correction bins \\

\bottomrule
\end{tabularx}
\end{table}

\end{document}